%% file: main.tex
\documentclass[10pt,twocolumn,letterpaper]{article}

\usepackage[pagenumbers]{wacv} 


\definecolor{wacvblue}{rgb}{0.21,0.49,0.74}
\usepackage[pagebackref,breaklinks,colorlinks,allcolors=wacvblue]{hyperref}
\usepackage{multirow}

\def\confName{WACV}
\def\confYear{2027}

\title{Depth-Only Open-Vocabulary 3D Semantic Segmentation For Privacy-Preserving Robotic Applications}

\author{
Xuying Huang \and Sicong Pan \and Maren Bennewitz
}

\begin{document}
\maketitle

{%
\renewcommand\thefootnote{}%
\footnotetext{%
X. Huang, S. Pan, and M. Bennewitz are with the Humanoid Robots Lab, University of Bonn, the Lamarr Institute for Machine Learning and Artificial Intelligence, and the Center for Robotics, Bonn, Germany.
This work has been partially funded by the German Federal Ministry of Research, Technology and Space~(BMFTR) under grant No.~16KIS1949 and the Robotics Institute Germany~(RIG). Corresponding author: huang@cs.uni-bonn.de}%
}%
\setcounter{footnote}{0}

\input{sec/0_abstract}    
\input{sec/1_intro}

\input{sec/2_related_work}
\input{sec/3_method}
\input{sec/4_experiment}
\input{sec/5_conclusion}

{
    \small
    \bibliographystyle{ieeenat_fullname}
    \bibliography{main}
}

\end{document}

%% file: sec/0_abstract.tex
\begin{abstract}
Privacy-preserving perception is increasingly important for robotic systems operating in real-world indoor environments, yet it remains underexplored in open-vocabulary 3D semantic segmentation.
We study this problem under an RGB-prohibited deployment setting motivated by scene-specific visual information disclosure, where real RGB observations are unavailable during scene acquisition and fusion.
To reflect this deployment constraint on existing 3D datasets, we adopt a stricter depth-only evaluation protocol that re-runs scene fusion without RGB and exposes only the resulting depth-derived geometry to the segmentation pipeline.
This constraint removes appearance cues that are often critical for open-vocabulary recognition, making depth-only predictions more uncertain and less reliable.
To address this challenge, we propose UTTO, a model-agnostic uncertainty-guided test-time optimization framework that uses structured predictive uncertainty as a reliability signal to refine predictions from frozen open-vocabulary 3D backbones.
Experiments across ScanNet and Matterport3D demonstrate consistent improvements over multiple depth-only backbones.
Privacy recoverability analyses and a real-robot semantic goal grounding case study further support the proposed privacy-constrained setting and applicability.

\end{abstract}

%% file: sec/1_intro.tex
\section{Introduction}
\label{sec:intro}

\begin{figure*}[t]
\centering
\includegraphics[width=0.8\textwidth]{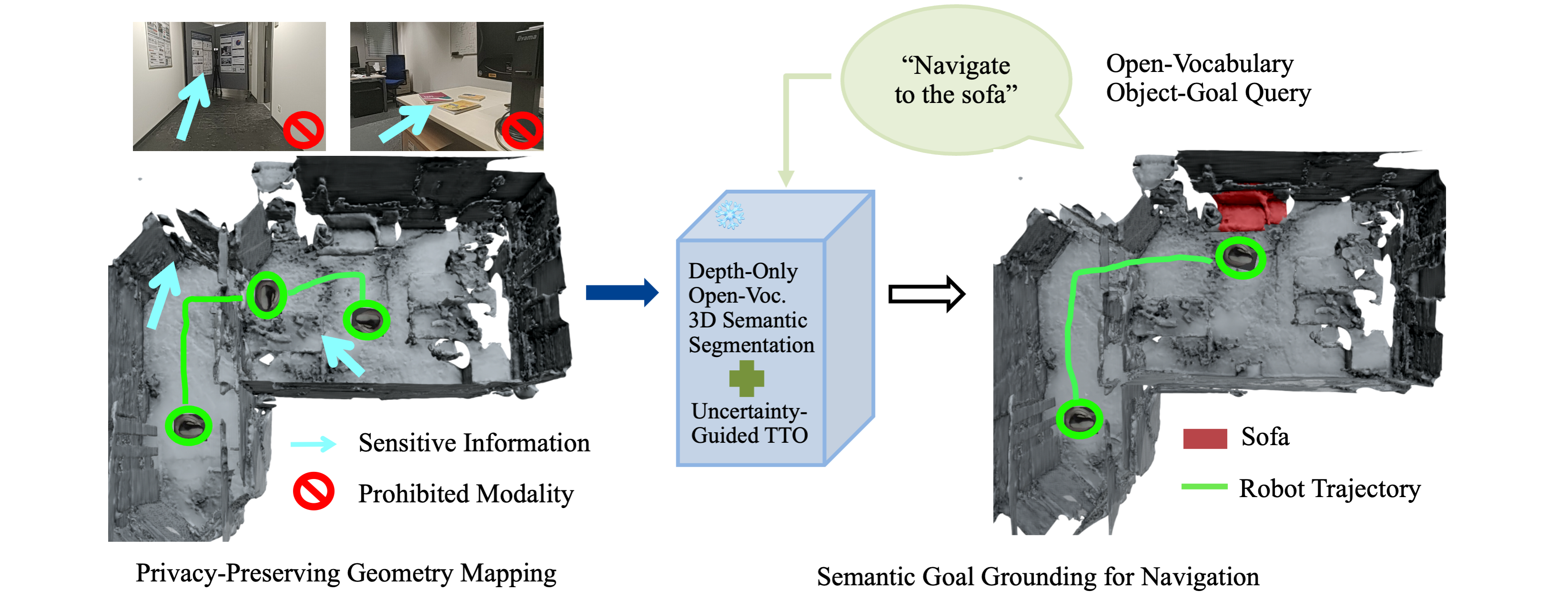}
\caption{
An example application scenario for privacy-preserving depth-only open-vocabulary 3D perception.
\textbf{Left:} During indoor mapping, RGB observations may reveal privacy-sensitive content such as screens, documents, or personal belongings; they are therefore treated as a prohibited modality and are not used by the perception pipeline.
The system therefore retains only depth-derived geometry, resulting in a geometry-only 3D map from the test scene.
\textbf{Right:} Given an open-vocabulary object-goal query such as ``navigate to the sofa'', UTTO~(Uncertainty-Guided Test-Time Optimization) performs depth-only open-vocabulary 3D segmentation on the geometry-only map and grounds the queried semantic region as a robot navigation goal.
}
\label{fig:teaser}
\vspace{-5mm}
\end{figure*}

Privacy-preserving perception has received growing attention as intelligent systems are increasingly deployed in real-world environments~\cite{huangarxiv}.
This requirement is particularly important for indoor robotic applications, where embodied systems and robots are expected to operate in homes, offices, hospitals, and other privacy-sensitive spaces.
However, the RGB observations commonly used by these systems may capture sensitive visual information such as human faces, computer screens, documents, and personal belongings, raising privacy concerns~\cite{huang26roman}.

Open-vocabulary 3D semantic segmentation is a key capability for general-purpose 3D scene understanding, enabling dense semantic parsing of unseen environments beyond a fixed label set.
This flexibility is important for downstream applications such as robotic navigation and human-scene interaction~\cite{gu24icra, werby24icra}, where the set of relevant categories may not be known in advance.
However, privacy-preserving perception remains underexplored in this setting.
Existing methods typically obtain rich semantic cues from RGB information~\cite{lee25cvpr, zhu24eccv, takmaz25ral}.
Although such cues have advanced open-vocabulary 3D recognition, they also make semantic inference dependent on RGB observations.
This dependence conflicts with deployments in which access to appearance observations must be restricted.

We therefore consider a privacy threat model focused on scene-specific appearance information revealed by RGB sensing, particularly readable text and personal appearance cues.
Under this threat model, depth-only 3D geometry provides a natural sensing modality for limiting appearance-level exposure~\cite{peng23cvpr, samet26dv}.
Accordingly, we define the setting studied in this paper as an RGB-prohibited deployment regime, where the robot relies only on depth observations and depth-derived geometry, while real RGB images from the test scene are unavailable to the perception pipeline.

Importantly, our deployment requirement is stricter than simply evaluating a segmentation model on a RGB-free 3D representation.
Conventional geometry-only evaluation specifies the evidence available to the segmentation model, but does not necessarily constrain whether RGB observations were available earlier during scene acquisition or reconstruction~\cite{peng23cvpr}.
In our setting, RGB is explicitly prohibited from the sensing stage onward, and scene fusion is performed using depth-derived geometry only.
For existing datasets, we enforce this information boundary by re-running scene fusion without RGB and exposing only the geometry-only representation to the segmentation pipeline.

This RGB-prohibited setting removes the texture and color cues that often provide critical semantic evidence for open-vocabulary recognition, making depth-only predictions more uncertain and less reliable.
Importantly, this uncertainty is not uniformly distributed across the scene.
Some regions remain stable because their geometry provides sufficient semantic evidence, while others become ambiguous due to missing appearance cues.
This structured uncertainty provides a useful signal for inference-time refinement: reliable predictions should be preserved, whereas uncertain predictions should be allowed to change under geometric and semantic regularization.
Since RGB observations are unavailable at deployment and we do not assume additional labeled data, such refinement should be performed at test time without updating the underlying model.

We therefore propose UTTO, a model-agnostic uncertainty-guided test-time optimization framework for depth-only open-vocabulary 3D semantic segmentation.
Starting from a frozen open-vocabulary 3D model, UTTO refines its scene predictions at inference time, making it applicable to different 3D backbones without retraining.
It constructs multiple predictions through label-preserving test-time perturbations and estimates vertex-level uncertainty from their agreement.
The resulting reliability guides the optimization to preserve stable predictions while selectively refining uncertain ones.
To further improve open-vocabulary recognition, UTTO combines scene-adaptive class prototypes, prompt-ensemble text prototypes for reducing text-side bias, and coordinate- and feature-space consistency for exploiting complementary geometric and semantic cues.

Fig.~\ref{fig:teaser} illustrates this privacy-preserving setting from a robotic application perspective and shows how UTTO supports downstream semantic goal navigation from geometry-only input.
Our contributions are as follows:
\begin{itemize}
\item We study depth-only open-vocabulary 3D semantic segmentation for privacy-preserving robotic applications under an RGB-prohibited deployment setting, and demonstrate its practical relevance through a real-robot semantic goal grounding case study.
\item We enforce a stricter depth-only evaluation protocol on existing 3D datasets by re-running scene fusion without RGB and exposing only depth-derived geometry to the segmentation pipeline.
\item We propose UTTO, a model-agnostic uncertainty-guided test-time optimization framework that exploits structured test-time uncertainty as a reliability signal to selectively refine predictions.
\end{itemize}

%% file: sec/2_related_work.tex
\section{Related Work}
\label{sec:RW}

\subsection{Open-Vocabulary 3D Semantic Segmentation}

Open-vocabulary 3D semantic segmentation is an important step toward general-purpose 3D scene understanding, as it enables models to parse 3D environments beyond a fixed set of predefined categories~\cite{wang25cvpr}.
Such an ability is crucial for downstream applications including robotic navigation, AR/VR, and human-scene interaction.

\textbf{RGB-Dependent Open-Vocabulary 3D Segmentation.}
Existing open-vocabulary 3D semantic segmentation methods commonly obtain semantic cues from appearance observations available at inference time.
One representative strategy is to extract dense features from posed RGB or RGB-D images using 2D open-vocabulary models and then lift these features to 3D points through camera projection and multi-view fusion.
OpenScene~\cite{peng23cvpr} is a representative example: its 2D fusion and 2D-3D ensemble settings back-project per-pixel open-vocabulary features to 3D points and combine them with 3D representations for semantic prediction.
Recent works further improve RGB-to-3D transfer using structured 2D priors.
Diff2Scene~\cite{zhu24eccv} leverages text-to-image diffusion representations and 2D mask embeddings as semantic queries for open-vocabulary 3D semantic segmentation.
XMask3D~\cite{wang24neurips} introduces cross-modal mask reasoning to align 3D representations with a 2D-text embedding space at the mask level, improving fine-grained segmentation boundaries.

Another line of work directly operates on colored 3D inputs.
Instead of explicitly lifting dense image features at inference time, these methods consume point clouds whose points are associated with RGB values.
For example, Mosaic3D~\cite{lee25cvpr} builds a foundation model for open-vocabulary 3D semantic and instance segmentation by exploiting colored 3D inputs.
Similarly, language-driven 3D scene understanding frameworks such as PLA~\cite{ding23cvpr} are applied to colored point-cloud inputs, where RGB channels provide additional appearance cues for point-level recognition.
Although these RGB-dependent methods have significantly advanced open-vocabulary 3D semantic segmentation, their inference pipelines require real RGB images, point-level colors, or image-derived features from the test scene.
This dependence may conflict with privacy-sensitive applications where RGB observations can expose faces, screens, documents, or personal belongings.

\textbf{RGB-Free Open-Vocabulary 3D Segmentation.}
RGB-free downstream inference provides a more privacy-compatible alternative by avoiding RGB appearance cues as input to the segmentation model.
Several recent works have explored pure-geometry inference for open-vocabulary 3D scene understanding.
OpenScene3D~\cite{peng23cvpr} introduces a 3D-distilled setting, where knowledge from 2D open-vocabulary models is distilled into a 3D network that predicts dense point features from 3D point positions at inference time.
Building on this direction, GGSD~\cite{wang24eccv} improves distilled 3D representations by incorporating geometric priors and self-distillation to better exploit the representational advantages of 3D point clouds.
OpenUrban3D~\cite{wang26tgrs} studies annotation-free open-vocabulary semantic segmentation for large-scale urban point clouds and generates semantic features directly from raw point clouds without requiring aligned multi-view images at inference time.
These works show that open-vocabulary 3D segmentation can be performed without online RGB feature fusion, making them more compatible with privacy-preserving deployment.

However, RGB-free open-vocabulary 3D segmentation remains harder because geometry alone often lacks the texture and color cues needed to distinguish geometrically similar categories.
As a result, predictions under depth-only inference can become spatially inconsistent and unreliable.
In contrast to existing RGB-free methods that directly use the frozen model predictions, we treat this reliability variation as an optimization signal and refine depth-only open-vocabulary predictions at test time without using real RGB observations, additional training, or ground-truth labels.

\begin{figure*}[t]
\centering
\includegraphics[width=0.7\textwidth]{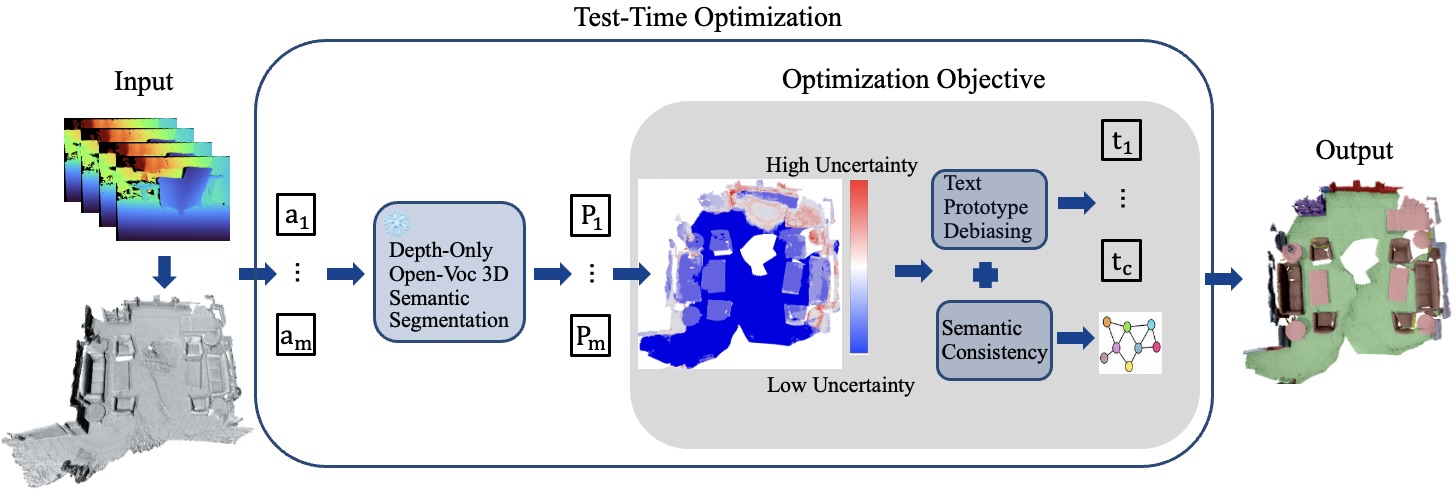}
\caption{
Overview of uncertainty-guided test-time optimization.
Given a depth-only geometry input, label-preserving test-time augmentations
$a_1,\ldots,a_M$ produce multiple predictions $P_1,\ldots,P_M$ from a frozen
depth-only open-vocabulary 3D segmentation backbone. Prediction agreement
provides a vertex-level reliability signal that identifies reliable scene
evidence for inference-time optimization, preserving stable predictions while
allowing uncertain regions to be refined. The optimization further incorporates
label-free prompt-ensemble CLIP text prototypes to reduce text-side bias, together
with geometric and feature-space semantic consistency. 
}
\label{fig:pipeline}
\vspace{-5mm}
\end{figure*}

\subsection{Uncertainty-Guided Refinement and Test-Time Optimization}

\textbf{Uncertainty-Guided Refinement.}
Uncertainty estimation is important for 3D semantic segmentation, where point-level predictions can be unreliable due to sparse observations, occlusion, and geometric ambiguity.
Existing works have exploited uncertainty in different ways.
In closed-set LiDAR or point-cloud segmentation, uncertainty is often used to assess prediction reliability and improve robustness or calibration~\cite{cortinhal20arxiv}.
In label-efficient 3D segmentation, it is commonly used for pseudo-label generation, pseudo-label filtering, or active sample selection~\cite{hu22eccv, xu23iccv}.
Recent open-world and open-vocabulary 3D methods further use uncertainty to identify unknown regions or to balance heterogeneous semantic cues from multiple foundation models~\cite{cen22eccv, li25cvpr}.
However, uncertainty remains largely unexplored as an inference-time reliability signal for depth-only open-vocabulary 3D semantic segmentation.
This setting is particularly challenging because RGB appearance cues are unavailable by design, making semantic predictions more ambiguous and spatially uneven in reliability.

\textbf{Test-Time Optimization.}
Test-time optimization and inference-time refinement aim to improve scene-level predictions without collecting labeled target data.
In open-vocabulary 3D segmentation, existing methods often refine 3D masks by leveraging class-agnostic 3D proposals, multi-view vision-language features, or mask-level feature aggregation~\cite{takmaz23neurips, nguyen24cvpr}.
Other training-free methods use foundation models to refine 3D prompts, merge superpoints, or aggregate open-world tags for open-vocabulary 3D scene understanding~\cite{xu25dv, tai24arxiv}.
These methods demonstrate the value of inference-time refinement, but they are mostly built on RGB-derived observations, image-level foundation-model features, or mask proposals that depend on appearance-rich inputs.
In contrast, UTTO performs uncertainty-guided test-time optimization directly on depth-only geometry-based predictions.
It optimizes per-point class distributions from frozen open-vocabulary 3D backbones, preserving reliable regions while refining uncertain ones under geometric and foundation-model priors, without using real RGB observations from the test scene.

%% file: sec/3_method.tex
\section{Our Method}
We propose an uncertainty-guided test-time optimization framework for depth-only open-vocabulary 3D semantic segmentation. 
Given a geometry-only scene, we keep all backbone and foundation encoders frozen and optimize the scene predictions at inference time.

As illustrated in Fig.~\ref{fig:pipeline}, our framework consists of three components.
First, label-preserving test-time perturbations provide multiple geometric observations, from which we obtain scene features and predictions. We then estimate vertex-level uncertainty and use the resulting reliability to construct scene-adaptive class prototypes from confident vertices.
Second, we reduce text-side bias in the unary predictions by updating each class's text prototype toward a scene-adaptive prototype estimated from its most reliable vertices, and re-classify all vertices with the debiased prototypes. 
Third, coordinate-space and feature-space consistency 
improve the performance by exploiting complementary geometric and semantic feature cues.

\subsection{Uncertainty-Guided Inference Optimization} \label{sec:Uncertainty}
Under the depth-only privacy-constrained setting, we leverage predictive uncertainty induced by label-preserving test-time perturbations as a label-free reliability signal, and use this uncertainty to identify reliable scene evidence for inference-time optimization.

To obtain multiple predictions without supervision or model training, we perform test-time augmentation on the input scene. 
Let \(\mathcal{A}=\{a_m\}_{m=1}^{M}\) denote the set of semantics-preserving augmentations. We use \(M\) augmented views composed of yaw rotation, isotropic scaling or axis flipping. 
These augmentations preserve scene semantics while inducing complementary prediction variation from the frozen model.

Given a geometry-only scene \(\mathcal{S}\) with \(V\) vertices and \(C\) semantic categories, the frozen open-vocabulary 3D model produces a per-vertex class distribution for each augmented input:
$
P_{(m)} = f_{\theta}(a_m(\mathcal{S}); T) \in \mathbb{R}^{V\times C},
\ m=1,\ldots,M,
$
where \(f_{\theta}\) denotes the frozen 3D model, and \(T\) denotes the text-prototype classifier, which will be described in Sec.~\ref{sec:text_classifier}. 
Each prediction is mapped back to the original vertex space before aggregation.
We additionally retain the corresponding per-vertex embeddings \(H^{(m)} \in \mathbb{R}^{V \times d}\), which are used for subsequent prototype construction.
In particular, we obtain the per-vertex embedding
$
\bar{H}
=
\frac{1}{M}
\sum_{m=1}^{M}
H^{(m)}.
$
Using the text-prototype classifier \(T\), the aggregated features produce the unary observations
\(Y=\{y_v\}_{v=1}^{V}\), where \(y_v\in\Delta^{C-1}\) denotes the class distribution of vertex \(v\).

We then estimate vertex-level reliability from the test-time predictions.
Let \(\rho_v\in[0,1]\) denote the resulting reliability score of vertex \(v\), where a larger \(\rho_v\) indicates a more confident prediction, while a smaller \(\rho_v\) indicates greater predictive uncertainty.
We define the corresponding uncertainty as
$
u_v=1-\rho_v.
$
We convert the reliability score into a vertex-wise reliability weight
$
w_v = \mathcal{W}(\rho_v).
$

Let
$
X=\{x_v\}_{v=1}^{V},
\ 
x_v\in\Delta^{C-1}
$
denote the per-vertex class distributions optimized at inference time.
We denote by
\(E_{\mathrm{data}}(X)\)
the uncertainty-guided semantic data term constructed from the aggregated observations, reliability weights, and prototype evidence described in Sec.~\ref{sec:text_classifier}.
In addition, we use
\(E_{\mathrm{cons}}(X)\)
to denote the coordinate- and feature-space consistency term described in Sec.~\ref{sec:feature_consistency}.

\subsection{Label-Free Text Prototype Debiasing}
\label{sec:text_classifier}

The semantic data term
\(E_{\mathrm{data}}\)
in Sec.~\ref{sec:Uncertainty} depends on the text-prototype classifier \(T\) used by the frozen open-vocabulary model.
We therefore construct a label-free text classifier by replacing single-template class prototypes with prompt-ensemble prototypes.
For each test-time prediction in Sec.~\ref{sec:Uncertainty}, the frozen 3D backbone produces a per-vertex feature
\(h_v^{(m)}\in\mathbb{R}^{d}\).
Open-vocabulary classification is performed by comparing this feature with a set of text prototypes.
Let
\(t_c\in\mathbb{R}^{d}\)
denote the text prototype of category \(c\).
The class logit is computed by cosine similarity:
$
z_{v,c}^{(m)}
=
\left\langle
\hat{h}_v^{(m)},
\hat{t}_c
\right\rangle,
$
where
\(\hat{h}_v^{(m)}\)
and
\(\hat{t}_c\)
are the \(\ell_2\)-normalized features.
The probability prediction
\(P_{(m)}\)
used in Sec.~\ref{sec:Uncertainty} is obtained by applying a softmax over the class logits.

A common strategy is to build each class prototype from a single template, such as
``a \{desk\} in a scene''.
However, a single prompt provides only one linguistic description of a category and can introduce wording-specific bias.
Since the 3D model is frozen, such bias directly affects the open-vocabulary decision boundary and propagates to all test-time predictions.
This issue is important in the depth-only setting, where appearance information is unavailable and recognition depends more strongly on the alignment between geometric features and language prototypes.

To mitigate prompt-induced bias without labels, we construct each text prototype using a set of generic templates~\citep{radford21icml}.
Let
\(\mathcal{Q}=\{q_k\}_{k=1}^{K}\)
be the template set and
\(n_c\)
be the name of category \(c\).
For each category, we generate
$
s_{c,k}=q_k(n_c),
\ 
k=1,\ldots,K
$
and encode the resulting prompts using the frozen CLIP~\citep{radford21icml} text encoder.
The class prototype is obtained by averaging normalized text embeddings followed by another normalization:
$
t_c
=
\operatorname{norm}
\left(
\frac{1}{K}
\sum_{k=1}^{K}
\operatorname{norm}
\left(
g_{\phi}(s_{c,k})
\right)
\right),
$
where
\(
\operatorname{norm}(x)=x/\|x\|_2
\).
The resulting text-prototype classifier is
\(
T=[t_1,\ldots,t_C]^{\top}
\).

During inference optimization, the text prototype provides a category-level language prior, while the reliability weight \(w_v\) from Sec.~\ref{sec:Uncertainty} determines the contribution of scene-specific feature evidence.
For each semantic category \(c\), let
\(\mathcal{V}_c\)
denote the vertices provisionally associated with that category.
We summarize their reliability-weighted scene evidence as
$
\mu_c
=
\operatorname{norm}
\left(
\frac{
\sum_{v\in\mathcal{V}_c}
w_v\bar{h}_v
}{
\sum_{v\in\mathcal{V}_c}
w_v
}
\right).
$
We combine the scene evidence \(\mu_c\) with the text prototype \(t_c\) to obtain the test-time class representation
$
\tilde{t}_c
=
\operatorname{norm}
\left(
(1-\eta)t_c+\eta\mu_c
\right),
$
where \(\eta\) controls the contribution of scene-specific evidence.
The resulting class representations form the scene-adaptive classifier
\(
\tilde{T}=[\tilde{t}_1,\ldots,\tilde{t}_C]^\top
\),
which produces the corresponding per-vertex unary observations
\(
\tilde{Y}=\{\tilde{y}_v\}_{v=1}^{V}
\).

We define the semantic data term as
$
E_{\mathrm{data}}(X)
=
\sum_{v=1}^{V}w_v
\left\|
x_v-\tilde{y}_v
\right\|_2^2.
$
The same prompt set and prototype construction procedure are applied to all categories.
No ground-truth labels or text-prototype training is used.
Consequently, the semantic evidence used by the inference solver remains fully label-free and scene-specific while preserving the open-vocabulary language prior.

\begin{table*}[t]
\centering
\scriptsize
\setlength{\tabcolsep}{6.0pt}
\caption{
Results on ScanNet20, ScanNet40, ScanNet200, and Matterport3D under non-private RGB-D and privacy-preserving RGB-prohibited settings.
The Input column specifies the evidence provided to each segmentation pipeline.
Pseudo-RGB denotes RGB views generated from depth observations, whereas depth-only geometry methods operate directly on depth-derived 3D geometry.
Bold numbers indicate the best result for each metric.
}

\vspace{-2mm}

\label{tab:scannet_results}
\resizebox{\textwidth}{!}{
\begin{tabular}{lllcccccccc}
\toprule
\multirow{2}{*}{Setting}
& \multirow{2}{*}{Input}
& \multirow{2}{*}{Method}
& \multicolumn{2}{c}{ScanNet20}
& \multicolumn{2}{c}{ScanNet40}
& \multicolumn{2}{c}{ScanNet200}
& \multicolumn{2}{c}{Matterport3D} \\
\cmidrule(lr){4-5}
\cmidrule(lr){6-7}
\cmidrule(lr){8-9}
\cmidrule(lr){10-11}
& &
& mIoU & mAcc
& mIoU & mAcc
& mIoU & mAcc
& mIoU & mAcc \\
\midrule

\multirow{2}{*}{Non-private}
& \multirow{2}{*}{RGB-D}
& Mosaic3D
& 51.5 & \textbf{78.3}
& 35.4 & \textbf{61.7}
& 14.3 & \textbf{29.0}
& 38.2 & \textbf{72.7} \\

&
& Mosaic3D + UTTO
& 55.0 & 77.4
& \textbf{38.1} & 60.7
& \textbf{14.7} & 28.8
& 41.0 & 71.9 \\

\midrule

\multirow{7}{*}{Privacy-preserving}
& Pseudo-RGB + Depth
& D2RGB + OpenSeg Fusion
& 22.2 & 40.5
& 11.9 & 22.7
& 1.4 & 3.9
& 5.1 & 7.0 \\

\addlinespace[1.5mm]

&
\multirow{6}{*}{Depth-only geometry}
& Mosaic3D-DepthOnly
& 31.6 & 54.1
& 19.1 & 32.9
& 4.5 & 8.7
& 17.9 & 37.2 \\

&
& \textbf{Mosaic3D-DepthOnly + UTTO (Ours)}
& 36.8 & 57.8
& 22.4 & 37.9
& 6.1 & 11.4
& 20.2 & 38.8 \\

&
& OpenScene3D
& 44.2 & 64.0
& 28.2 & 43.3
& 4.7 & 8.8
& 31.8 & 44.1 \\

&
& \textbf{OpenScene3D + UTTO (Ours)}
& 47.3 & 66.8
& 29.7 & 45.2
& 5.9 & 10.2
& 33.2 & 46.0 \\

&
& SAS
& 58.9 & 68.8
& 27.4 & 36.6
& 5.4 & 7.6
& 39.8 & 49.2 \\

&
& \textbf{SAS + UTTO (Ours)}
& \textbf{61.2} & 72.5
& 28.9 & 38.2
& 6.2 & 8.1
& \textbf{41.1} & 51.0 \\

\bottomrule
\end{tabular}
}
\vspace{-2mm}
\end{table*}

\begin{figure*}[t]
\centering
\includegraphics[width=\textwidth]{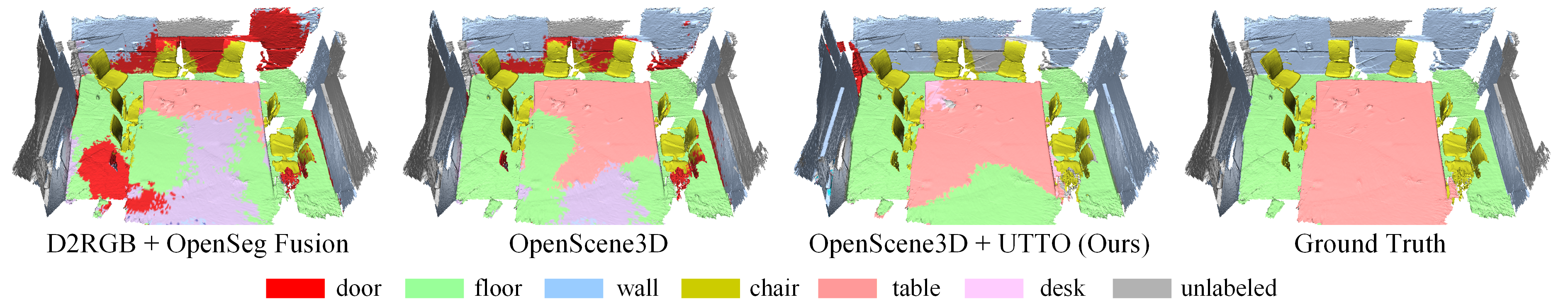}
\caption{
Qualitative comparison under the privacy-preserving depth-only setting.
All methods use the same semantic color palette, with the legend showing the main classes in the scene.
}
\label{fig:qualitative_comparison}
\vspace{-5mm}
\end{figure*}

\subsection{Geometry and Feature-Space Semantic Consistency}
\label{sec:feature_consistency}
The uncertainty-guided semantic term determines how the predictions and prototype evidence contribute to the inference optimization.
However, vertices are not independent: spatially neighboring points often share semantic labels, and vertices with similar geometric or learned representations should remain consistent even when local predictions are uncertain.
We therefore complement the semantic evidence with coordinate-space and feature-space consistency constraints.
We first construct a geometric relation graph
\(\mathcal{E}_g\)
from the local 3D structure of the scene.
The geometric relations capture spatial and structural consistency between nearby vertices and suppress isolated noisy predictions.
We formulate the corresponding geometric consistency term as
$
E_{\mathrm{geo}}(X)
=
\sum_{(i,j)\in\mathcal{E}_g}
S^{g}_{ij}
\left\|
x_i-x_j
\right\|_2^2,
$
where
\(S^{g}_{ij}\)
denotes the geometric affinity between vertices \(i\) and \(j\).

Coordinate-space consistency alone, however, does not always imply semantic consistency, especially in the depth-only privacy-preserving setting where RGB appearance cues are unavailable.
To complement the geometric relations, we further establish semantic relations in a learned feature space.
To preserve the depth-only setting, we render depth-derived geometric views by colorizing depth values with a fixed colormap rather than using real RGB appearance.
These colorized depth renderings encode only geometric depth variation and do not contain scene texture, point-level color, or real image information.
We extract features from these RGB-formatted depth renderings using a frozen DINO encoder~\citep{oquab23arxiv} and back-project them to the corresponding 3D vertices, obtaining a feature
\(\phi_v\)
for each vertex.
The resulting representations are used to establish feature-space semantic relations
\(\mathcal{E}_s\),
with
\(S^{s}_{ij}\)
denoting the semantic affinity between vertices \(i\) and \(j\).
The corresponding feature-space consistency term is formulated as
$
E_{\mathrm{sem}}(X)
=
\sum_{(i,j)\in\mathcal{E}_s}
S^{s}_{ij}
\left\|
x_i-x_j
\right\|_2^2.
$

The geometric and feature-space consistency terms together define
$
E_{\mathrm{cons}}(X) = \lambda_g E_{\mathrm{geo}}(X) + \lambda_s E_{\mathrm{sem}}(X),
$
where \(\lambda_g\) and \(\lambda_s\) control the contributions of geometric and feature-space consistency, respectively.
Together with the data term
\(E_{\mathrm{data}}\),
the final inference objective is
\begin{equation}
\small
X^{\star}
=
\arg\min_{X:\,x_v\in\Delta^{C-1}}
\left[
E_{\mathrm{data}}(X)
+
E_{\mathrm{cons}}(X)
\right]
\label{eq:final_objective}
\end{equation}
The final semantic label of vertex \(v\) is obtained by
$
\hat{y}_v
=
\arg\max_c
x^{\star}_{v,c}.
$

%% file: sec/4_experiment.tex
\section{Experiments}

Our experiments address three main questions.
First, we instantiate the target RGB-prohibited deployment setting on existing 3D datasets and evaluate whether UTTO consistently improves depth-only open-vocabulary 3D segmentation across different backbones and datasets.
Second, we examine the privacy implications by measuring the recoverability of sensitive textual and appearance information from depth observations and depth-to-RGB reconstructions.
Third, we demonstrate the practical relevance of the proposed setting through a real-robot privacy-preserving semantic goal grounding case study.

\subsection{Depth-Only Open-Vocabulary 3D Segmentation under RGB-Prohibited Evaluation}

\subsubsection{RGB-Prohibited Evaluation Setup}

\textbf{Datasets.}
We evaluate on ScanNet~\citep{dai17cvpr} and Matterport3D~\citep{chang17_3dv}, two widely used indoor RGB-D datasets with dense 3D semantic annotations.
For ScanNet, we report results on the validation sets of ScanNet20, ScanNet40, and ScanNet200~\citep{rozenberszki22eccv}, where ScanNet40 follows the 40-class NYUv2~\citep{silberman12eccv} label taxonomy.
For Matterport3D, we adopt the 21-class semantic label set.

\textbf{RGB-Prohibited Scene Reconstruction.}
To instantiate the target deployment constraint on these RGB-D datasets, we re-run scene reconstruction using depth observations only.
RGB is excluded from pose estimation and scene fusion, and the resulting geometry-only reconstruction is used as the sole scene representation for downstream segmentation.
To improve reconstruction accuracy without relying on RGB, we use a depth-only pipeline that combines frame-to-model tracking~\citep{newcombe11ismar}, geometry-only submap registration~\citep{choi15cvpr}, global pose optimization, and GT-blind geometric consistency refinement.
The optimized trajectory is then used to re-integrate the full depth sequence into a TSDF~\citep{curless96siggraph} mesh with geometry-preserving surface processing.
This differs from conventional geometry-only evaluation that may remove color only after an RGB-assisted reconstruction.
Detailed reconstruction and fusion procedures are provided in the supplementary material.

\textbf{Evaluation and Metrics.}
For all open-vocabulary methods, target semantic categories are specified only by text labels at inference time, without task-specific supervised training on the evaluated label set.
Because our depth-only reconstructions do not share the same vertex set as the annotated reference meshes, we transfer semantic predictions from the reconstructed geometry to the reference mesh vertices following~\citep{dai17cvpr}.
The aligned predictions are compared against the ground-truth semantic annotations, and we report mean Intersection over Union (mIoU) and mean class Accuracy (mAcc).

\textbf{Baselines.}
We compare methods under the two sensing settings: a non-private setting with access to real RGB-D observations and the target privacy-preserving setting in which real RGB observations are prohibited.
Under the non-private setting, we include Mosaic3D~\cite{lee25cvpr} with its standard RGB-D input as an open-vocabulary reference.
We also apply UTTO to its RGB-D predictions, yielding Mosaic3D+UTTO.
These results provide contextual references for assessing the effect of removing real appearance cues.

Under the privacy-preserving setting, we first include D2RGB+OpenSeg Fusion as an alternative that starts only from depth observations but reconstructs pseudo-RGB views using a depth-conditioned generative model~\cite{liu2025iccv}.
OpenSeg~\cite{ghiasi22eccv} is applied to the generated views, and the resulting 2D predictions are fused into 3D using the corresponding depth maps.
This baseline examines whether generated appearance can compensate for the absence of real RGB observations.

Our primary comparisons use depth-only geometry directly.
We evaluate Mosaic3D-DepthOnly~\cite{lee25cvpr}, OpenScene3D~\cite{peng23cvpr}, and SAS~\citep{li25iccv} under the same RGB-prohibited reconstruction protocol.
For Mosaic3D-DepthOnly, we replace the RGB channels with constant zero values; OpenScene3D~(OpenSeg-distilled 3D model) and SAS operate directly on the geometry-only reconstruction.
We apply UTTO to all three frozen backbones without task-specific supervised training or ground-truth labels at inference time.

UTTO implementation details and hyperparameter settings are provided in the supplementary material, and all experiments are conducted on a single NVIDIA A100 GPU.

\subsubsection{Depth-Only Open-Vocabulary 3D Segmentation}

\textbf{Depth-Only Segmentation Performance.}
As shown in Table~\ref{tab:scannet_results}, under the privacy-preserving depth-only geometry setting, UTTO consistently improves both mIoU and mAcc for all three backbones across all evaluated benchmarks.
These improvements hold across both ScanNet and Matterport3D and across increasingly fine-grained ScanNet label spaces, indicating that UTTO is not tied to a particular backbone, dataset, or category granularity.

\textbf{Effect of Removing RGB Cues.}
Within the same Mosaic3D backbone, replacing RGB-D input with depth-only geometry leads to a substantial performance drop across all benchmarks, providing a controlled indication that real RGB observations contribute important semantic evidence.
However, this gap is not uniform across architectures or label spaces.
SAS achieves particularly strong geometry-only performance on the coarser ScanNet20 setting, even surpassing the non-private Mosaic3D RGB-D reference, but this advantage does not persist on the more fine-grained ScanNet40 and ScanNet200 settings.
This suggests that strong geometry-based representations can compensate for missing appearance cues when coarse geometric structure is sufficiently discriminative, whereas finer-grained open-vocabulary recognition remains more challenging without RGB appearance information.

\textbf{Pseudo-RGB Comparison.}
The privacy-preserving D2RGB+OpenSeg Fusion baseline performs substantially worse than the optimized depth-only geometry methods.
Although this baseline also satisfies the depth-only sensing constraint, it attempts to recover missing appearance cues through generated pseudo-RGB views.
Its weaker performance suggests that, in this setting, generated appearance does not reliably recover the semantic benefit provided by real RGB cues.
In contrast, directly refining geometry-based predictions with uncertainty-guided test-time optimization yields consistently stronger performance without introducing real RGB observations.
Fig.~\ref{fig:qualitative_comparison} provides a qualitative comparison under the RGB-prohibited setting, showing that UTTO produces cleaner and more spatially consistent predictions than both pseudo-RGB fusion and the unoptimized geometry-only backbone.

\textbf{Diagnostic under RGB-D Input.}
On the non-private Mosaic3D RGB-D reference, UTTO improves mIoU across all evaluated benchmarks, indicating that the proposed optimization is not limited to depth-only representations.
However, these gains are accompanied by consistent decreases in mAcc, suggesting a different refinement trade-off when real RGB appearance cues are available.
In contrast, under the RGB-prohibited depth-only setting, UTTO consistently improves both mIoU and mAcc across all tested backbones and benchmarks.
These results indicate that UTTO can also extend beyond depth-only input, while yielding more consistent and balanced improvements when RGB appearance cues are unavailable.

\textbf{Supplementary Analyses.}
We provide two sets of additional experiments in the supplementary material.
First, we further position the RGB-prohibited depth-only setting through comparisons under relaxed sensing and supervision constraints and under the standard geometry-only benchmark protocol.
The former highlights the combined challenges of missing appearance cues and the absence of task-specific supervision, while the latter shows that UTTO remains effective but generally yields smaller gains than under our RGB-prohibited reconstruction, further distinguishing the two evaluation regimes.
Second, additional analyses of UTTO show that the method benefits from uncertainty guidance and its semantic-consistency components, remains effective with alternative feature representations and different numbers of test-time augmentations, and captures spatially structured uncertainty that serves as a useful reliability signal for refinement.

\begin{table}[t]
\centering
\scriptsize
\setlength{\tabcolsep}{3.0pt}

\caption{
Privacy-sensitive information recoverability analysis.
}

\vspace{-2mm}

\begin{subtable}{\columnwidth}
\centering
\caption{Sensitive Text Recoverability}
\label{tab:privacy_text}
\resizebox{\columnwidth}{!}{
\begin{tabular}{lcccc}
\toprule
Metric
& RGB
& Depth
& Matched D2RGB
& Shuffled D2RGB
\\
\midrule
Exact recovery
& 331/384
& 0/384
& 0/384
& 0/384
\\

Character similarity
& 100.0\%
& 0.0\%
& 24.9\%
& 23.7\%
\\
\bottomrule
\end{tabular}
}
\end{subtable}


\begin{subtable}{\columnwidth}
\centering
\caption{Appearance Attribute Recoverability}
\label{tab:privacy_appearance}
\resizebox{\columnwidth}{!}{
\begin{tabular}{lcccc}
\toprule
Metric
& RGB
& Depth
& Matched D2RGB
& Shuffled D2RGB
\\
\midrule
Macro accuracy
& 95.0\%
& 0.0\%
& 22.5\%
& 21.7\%
\\
\bottomrule
\end{tabular}
}
\end{subtable}

\label{tab:privacy_recoverability}
\vspace{-5mm}
\end{table}

\subsection{Privacy Recoverability Analysis}
\label{sec:privacy_recoverability}

\textbf{Privacy Threat Model and Scope.}
Following the LINDDUN privacy threat modeling framework~\citep{deng11re}, we focus on \emph{Data Disclosure}: whether privacy-sensitive information directly observable from RGB can still be recovered from the depth observations retained by our pipeline or from D2RGB reconstructions derived from them.
Accordingly, we consider an adversary that has access to the depth input or its D2RGB reconstruction, knows the reconstruction procedure, and can apply off-the-shelf recognition models to infer sensitive information, but does not have access to the original RGB image.
Our evaluation considers two representative forms of visual information disclosure exposed by RGB sensing: sensitive textual content and personal appearance attributes.
Although depth may still expose geometric or temporal cues for identity inference, such as person re-identification or gait recognition~\citep{wu17tip}, these risks fall outside the scope of our disclosure-oriented analysis.

\textbf{Privacy Dataset and Evaluation.}
To enable a targeted privacy analysis, we collect a small privacy dataset of 80 images spanning four privacy-sensitive categories: credit card, passport, private chat, and face, with 20 images per category.
Text recoverability is evaluated on the first three categories (60 images), while appearance recoverability is assessed on the passport and face subsets (40 images), with the passport subset shared by both analyses.
For each analysis, we compare RGB, Depth, Matched D2RGB reconstructions generated from the corresponding depth input, and Shuffled D2RGB reconstructions from unrelated samples.

\subsubsection{Sensitive Text Recoverability}
\label{sec:privacy_text}

We evaluate whether textual information readable from the original RGB observations remains recoverable from Depth or D2RGB.
We use \texttt{PP-OCRv6\_medium}~\citep{zhang2026arxiv} with a confidence threshold of 0.95 as the OCR backend.

\textbf{Exact Recovery.}
We first evaluate exact recovery, which measures whether a complete text string is correctly recovered.
Across the 60 text-containing images, the RGB OCR detects 384 high-confidence text candidates, since each image may contain multiple textual regions.
Among them, 331 are manually verified as fully correct, corresponding to an exact recovery rate of 86.2\%.
In contrast, no complete reference string is recovered from Depth, Matched D2RGB, or Shuffled D2RGB.
The 331 verified RGB-readable strings are then used as the reference set for the subsequent character-level analysis.

\textbf{Character Similarity.}
We next evaluate partial textual recovery over the 331 verified RGB-readable references using normalized Levenshtein similarity~\citep{li07tpami}.
Depth yields no measurable character-level recovery.
Matched D2RGB achieves 24.9\% character similarity, only 1.2 percentage points above the 23.7\% Shuffled D2RGB control.
This small matched--shuffled gap suggests that the reconstructed text-like content is dominated by generic generative priors rather than reliable recovery of sample-specific private text.

\subsubsection{Appearance Attribute Recoverability}
\label{sec:privacy_appearance}

We next examine whether Depth or D2RGB reveals privacy-sensitive appearance attributes.
We consider three attributes that are directly observable from RGB---hair color, iris color, and eyeglass presence~\citep{dantcheva10mmsp}.
Following~\citep{bai25arXiv}, we use Qwen3-VL-8B-Instruct as the visual recognition model and query the same attributes for all input modalities.
Each attribute is evaluated over the 40 passport and face images.
For each input modality, we compute accuracy separately for the three attributes and report their unweighted average as the macro accuracy.

As shown in Table~\ref{tab:privacy_appearance}, RGB retains strong appearance information, achieving a macro accuracy of 95.0\%, whereas Depth yields 0.0\% across all three attributes.
Matched D2RGB achieves 22.5\% macro accuracy, comparable to the 21.7\% obtained by Shuffled D2RGB.
The fact that the matched reconstruction performs similarly to the shuffled control indicates that the synthesized appearance is not reliably associated with the corresponding input sample.
This distinguishes plausible appearance synthesis from true appearance recovery.
Detailed privacy dataset and per-attribute results are provided in the supplementary material.

\begin{table}[t]
\centering
\caption{
Real-robot semantic goal grounding results.
}
\label{tab:real_robot_goal_grounding}

\vspace{-2mm}

\resizebox{\columnwidth}{!}{
\begin{tabular}{l|cccc|c|c}
\hline
Method
& Chair & Sofa & Desk & Door & Total & Hit Rate \\
\hline
OpenScene3D
& 1/1 & 1/1 & 2/5 & 1/4 & 5/11 & 45.5\% \\

OpenScene3D + UTTO (Ours)
& 2/3 & 1/1 & 2/3 & 1/1 & 6/8 & 75.0\% \\
\hline
\end{tabular}
}
\vspace{-5mm}
\end{table}

\subsection{Privacy-Preserving Semantic Goal Grounding Case Study}

To demonstrate downstream applicability, we evaluate class-level semantic goal grounding in the real-world indoor robot scene shown in Fig.~\ref{fig:teaser}.
The robot operates on a pre-reconstructed geometry-only 3D map and receives open-vocabulary queries such as chair, sofa, desk, and door, without using real RGB observations.
Predicted semantic regions are projected to a top-down map, where each retained connected component generates one candidate navigation goal pose; therefore, different predictions may produce different numbers of candidates.
We report the fraction of candidate poses that fall within valid goal regions derived from coarse class-level annotations.
As shown in Table~\ref{tab:real_robot_goal_grounding}, OpenScene3D+UTTO improves the overall candidate goal-pose hit rate from 45.5\% (5/11) to 75.0\% (6/8).
Detailed task definitions, evaluation protocols, and qualitative results are provided in the supplementary material.

%% file: sec/5_conclusion.tex
\section{Conclusion}

We study depth-only open-vocabulary 3D semantic segmentation under an RGB-prohibited setting for privacy-preserving robotic applications and propose UTTO, a model-agnostic uncertainty-guided test-time optimization framework.
We instantiate this setting on existing benchmarks by re-running scene reconstruction without RGB and evaluating only on the resulting depth-derived geometry.
Experiments across multiple backbones, privacy recoverability analyses, and a real-robot semantic goal grounding case study demonstrate the effectiveness and practical relevance of the proposed approach.
Future work will explore privacy-filtered sensing strategies, such as extremely low-resolution RGB, that retain additional semantic cues while limiting the exposure of privacy-sensitive information.